\documentclass[runningheads]{llncs}

\usepackage{tabularx}
\usepackage{xcolor}
\usepackage{graphicx,float,epstopdf}

\usepackage{url}
\usepackage{bbold}

\usepackage{algorithm}
\usepackage{algorithmic}

\usepackage{tikz}
\usepackage{comment}
\usepackage{color}
\usepackage{subcaption}
\usepackage{graphicx}
\usepackage{amsmath}
\usepackage{amssymb}
\usepackage{booktabs}
\usepackage{xcolor}
\usepackage{colortbl}
\usepackage{adjustbox}
\usepackage{dsfont}
\usepackage{multirow}

\definecolor{LightGray}{rgb}{0.9,0.9,0.9}
\definecolor{Gray}{gray}{0.95}

\newcommand{\real}{\mathbb{R}}
\newcommand{\xx}{\mathbf{x}}
\newcommand{\yy}{\mathbf{y}}

\newcommand{\qq}{\mathbf{q}}
\begin{document}\newcommand{\enzo}[1]{{\textcolor{red}{[#1]}}}

\newcommand\blfootnote[1]{%
  \begingroup
  \renewcommand\thefootnote{}\footnote{#1}%
  \addtocounter{footnote}{-1}%
  \endgroup
}

\title{Maximum Entropy on Erroneous Predictions:\\ Improving model calibration for medical image segmentation}

\author{Agostina J. Larrazabal$^{1,4}$, César Martínez$^1$, Jose Dolz$^{2,3*}$, Enzo Ferrante$^{1*}$}
\institute{$^1$Research institute for signals, systems and computational intelligence, sinc(i), FICH-UNL / CONICET, Santa Fe, Argentina\\
 $^2$LIVIA, ETS Montreal, Canada \\
 $^3$ International Laboratory on Learning Systems (ILLS)\\
 $^4$ Tryolabs, Uruguay}

\titlerunning{ }

\maketitle

\begin{abstract} 
Modern\blfootnote{$^*$J. Dolz and E. Ferrante contributed equally to this article.}  deep neural networks achieved remarkable progress in medical image segmentation tasks. However, it has recently been observed that they tend to produce overconfident estimates, even in situations of high uncertainty, leading to poorly calibrated and unreliable models. In this work we introduce Maximum Entropy on Erroneous Predictions (MEEP), a training strategy for segmentation networks which selectively penalizes overconfident predictions, focusing only on misclassified pixels. 
Our method is agnostic to the neural architecture, does not increase model complexity and can be coupled with multiple segmentation loss functions. 
We benchmark the proposed strategy in two challenging segmentation tasks: white matter hyperintensity lesions in magnetic resonance images (MRI) of the brain, and atrial segmentation in cardiac MRI. The experimental results demonstrate that coupling MEEP with standard segmentation losses leads to improvements not only in terms of model calibration, but also in segmentation quality.

\end{abstract}
\section{Introduction}

Deep learning have become the \textit{de facto} solution for medical image segmentation. Nevertheless, despite their ability to learn highly discriminative features, these models have shown to be poorly calibrated, often resulting in over-confident predictions, even when they are wrong \cite{guo2017calibration}. When a model is miscalibrated, there is little correlation between the confidence of its predictions and how accurate such predictions actually are \cite{karimi2020improving}. 
This results in a major problem, which can have catastrophic consequences in medical diagnosis systems where decisions may depend on predicted probabilities.
As shown in \cite{czolbe2021segmentation}, the uncertainty estimates inferred from segmentation models can provide insights into the confidence of any particular segmentation mask, and highlight areas of likely errors for the practitioner. In order to improve the accuracy and reliability of these models, it is crucial to develop both accurate and well-calibrated systems.
Despite the growing popularity of calibration for image classification \cite{guo2017calibration,liu2022devil,mukhoti2020calibrating}, the impact of miscalibrated networks on image segmentation, especially in the realm of biomedical images, has only recently begun to be explored \cite{mehrtash2020confidence}.

\noindent \textbf{Contribution.} In this work, we propose a novel method based on entropy maximization to enhance the quality of pixel-level segmentation posteriors.
Our %working 
hypothesis is that penalizing low entropy on the probability estimates for erroneous pixel predictions during training should help to avoid overconfident estimates in situations of high uncertainty. The underlying idea is that, if a pixel is difficult to classify, it is better assigning uniformly distributed (i.e. high entropy) probabilities to all classes, rather than being overconfident on the wrong class. To this end, we design two simple regularization terms which push the estimated posteriors for misclassified pixels towards a uniform distribution by penalizing low entropy predictions. We benchmark the proposed method in two challenging medical image segmentation tasks.

Last, we further show that assessing segmentation models only from a discriminative perspective does not provide a complete overview of the model performance, and argue that including calibration metrics should be preferred. This will allow to not only evaluate the segmentation power of a given model, but also its reliability, of pivotal importance in healthcare. 

\noindent \textbf{Related work.} Obtaining well-calibrated probability estimates of supervised machine learning approaches has attracted the attention of the research community even before the deep learning era, including approaches like histogram %binning
\cite{zadrozny2001obtaining} or Bayesian binning \cite{naeini2015obtaining}. Nevertheless, with the increase of popularity of deep neural networks, several works to directly address the calibration of these models have recently emerged. %A broad span of methods have been presented to quantify predictive uncertainty in deep neural networks.
For instance, Bayesian neural networks learn a posterior distribution over parameters that quantifies parameter uncertainty --a type of \textit{epistemic uncertainty}--, providing a natural approach to quantify model uncertainty. Among others, well-known Bayesian methods include variational inference \cite{blundell2015weight}, dropout-based variational inference \cite{gal2016dropout} or stochastic expectation propagation \cite{hernandez2015probabilistic}. 
A popular non-Bayesian method is ensemble learning, a simple strategy that improves both the robustness and calibration performance of predictive models \cite{lakshminarayanan2016simple,stickland2020diverse,wen2020batchensemble,larrazabal2021orthogonal}. 
However, even though this technique tends to improve the networks calibration, it does not directly promote uncertainty awareness. Furthermore, ensembling typically requires retraining several models from scratch, incurring into computationally expensive steps for large datasets and complex models. Guo et \textit{al.} \cite{guo2017calibration} empirically evaluated several post training ad-hoc calibration strategies, finding that a simple temperature scaling of logits yielded the best results. A drawback of this simple strategy, though, is that calibration performance largely degrades under data distribution shift \cite{ovadia2019can}.

Another alternative is to address the calibration problem during training, for example by clamping over-confident predictions. In \cite{pereyra2017regularizing}, authors proposed to regularize the neural network output by penalizing low entropy output distributions, which was achieved by integrating an entropy regularized term into the main learning objective. 
We want to emphasize that, even though the main motivation in \cite{pereyra2017regularizing} was to achieve better generalization by avoiding overfitting, recent observations \cite{muller2019does} highlight that these techniques have a favorable effect on model calibration. 
In a similar line of work, \cite{mukhoti2020calibrating} empirically justified the excellent performance of focal loss to learn well-calibrated models. More concretely, authors observed that focal loss \cite{lin2017focal} minimizes a Kullback-Leibler (KL) divergence between the predicted softmax distribution and the target distribution, while increasing the entropy of the predicted distribution.

An in-depth analysis of the calibration quality obtained by training segmentation networks with the two most commonly used loss functions, Dice coefficient and cross entropy, was conducted in \cite{mehrtash2020confidence}. In line with \cite{milletari2016v,sander2019towards}, authors showed that loss functions directly impact calibration quality and segmentation performance, noting that models trained with soft Dice loss tend to be poorly calibrated and overconfident. Authors also highlight the need to explore new loss functions to improve both segmentation and calibration quality. 
Label smoothing (LS) has also been proposed to improve calibration in segmentation models. 
Islam et al \cite{islam2021spatially} 
propose a label smoothing strategy for image segmentation by designing a weight matrix with a Gaussian kernel which is applied across the one-hot encoded expert labels to obtain soft class probabilities. 
They stress that the resulting label probabilities for each class are similar to one-hot within homogeneous areas and thus preserve high confidence in non-ambiguous regions, whereas uncertainty is captured near object boundaries. Our proposed method achieves the same effect but generalized to different sources of uncertainty by selectively maximizing the entropy only for difficult to classify pixels.

\section{Maximum Entropy on Erroneous Predictions}
\label{sec:methods}

Let us have a training dataset $\mathcal{D} = \{(\xx, \yy)_n\}_{1 \leq n \leq |\mathcal{D}|}$, where $\xx_n \in \real^{\Omega_n}$ denotes an input image and $\yy_n \in \{ 0,1 \}^{\Omega_n \times K}$ its corresponding pixel-wise one-hot label. $\Omega_n$ denotes the spatial image domain and $K$ the number of segmentation classes. 
We aim at training a model, parameterized by $\theta$, which approximates the underlying conditional distribution $p(\mathbf{y}|\mathbf{x},\theta)$, where $\theta$ is chosen to optimize a given loss function. The output of our model, at a given pixel $i$, is given as $\hat{y}_i$, whose associated class probability is $p(\mathbf{y}|\mathbf{x},\theta)$. Thus, $p(\hat{y}_{i,k} = k|x_i,\theta)$ will indicate the probability that a given pixel (or voxel) $i$ is assigned to the class $k \in K$. For simplicity, we will denote this probability as $\hat{p}_{i,k}$.

\begin{figure}[t!]
     \centering
     \includegraphics[width=\textwidth]{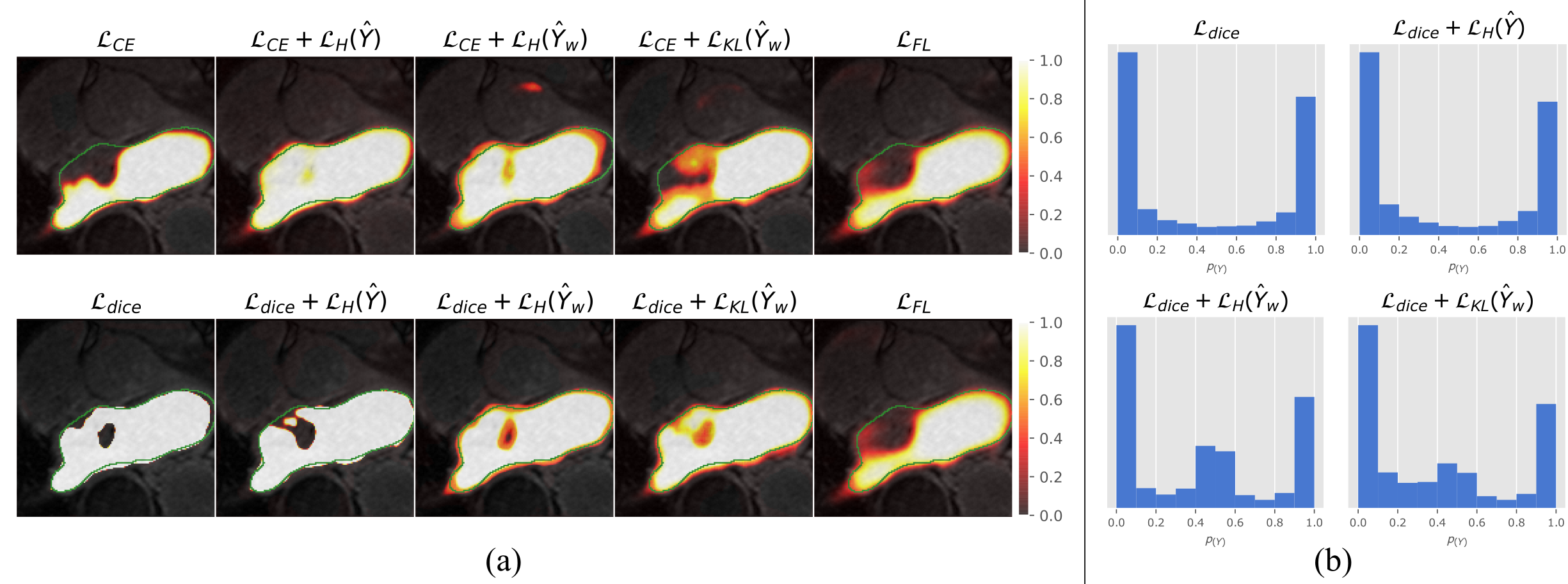}
\caption{(a) Results obtained on the left atrium (LA) segmentation task 
(ground truth in green). From left to right: \textbf{1st column} corresponds to UNet trained with standard segmentation losses, i.e., Dice ($\mathcal{L}_{dice}$) and cross entropy ($\mathcal{L}_{CE}$). \textbf{2nd column:} we include a regularization term $\mathcal{L}_{H}(\hat{Y})$,  which penalizes low entropy in all the predictions, 
\cite{pereyra2017regularizing}. \textbf{3rd and 4th columns:} results obtained with variations of the proposed MEEP method which penalizes low entropy predictions only on misclassified pixels. We can clearly observe how the proposed MEEP models push the predicted probabilities towards 0.5 in highly uncertain areas. \textbf{5th column:} results with focal loss. (b) Distribution of the magnitude of softmax probabilities on the Left Atrium dataset obtained when training with different loss functions. These plots motivate the proposed \textit{selective} confidence penalty (\textit{bottom row}) over prior work (\textit{top right}) \cite{pereyra2017regularizing}, as the resulting probability distributions produced by our method are smoother, leading to better calibrated networks.}
 \label{fig:softmax_prob}
    \label{fig:qualitative_LA}
\end{figure}

Since confident predictions correspond to low entropy output distributions, a network is overconfident when it places all the predicted probability on a single class for each training example, which is often a symptom of overfitting \cite{szegedy2016rethinking}. Therefore, maximizing the entropy of the output probability distribution encourages high uncertainty (or low confidence) in the network predictions. In contrast to prior work \cite{pereyra2017regularizing}, which penalizes low entropy in the entire output distributions, \textit{we propose to selectively penalize overconfidence exclusively for those pixels which are misclassified}, i.e. the more challenging ones. To motivate our strategy, we plot the distribution of the magnitude of softmax probabilities in Figure \ref{fig:softmax_prob}.b. It can be observed that for models trained with standard $\mathcal{L}_{dice}$ loss \cite{milletari2016v}, most of the predictions lie in the first or last bin of the histogram
We hypothesize that encouraging the network to assign high entropy values solely to erroneous predictions (i.e. uniformly distributed probabilities) will help to penalize overconfidence in complex scenarios. To this end, for every training iteration we define the set of misclassified pixels as $\hat{\yy}_w =\{y_i |\hat{y}_i\neq y_i\}$. We can then compute the entropy for this set as:

\begin{equation}\label{eq:neg_H}
\mathcal{H}(\hat{\yy}_w)=-\frac{1}{|\hat{\yy}_w|} \sum_{k, i\in \hat{\yy}_w } \hat{p}_{i,k}\log \hat{p}_{i,k},
\end{equation}

\noindent where $|\cdot|$ is used to denote the set cardinality. As we aim at maximizing the entropy of the output probabilities $\hat{\yy}_w$ (eq. \eqref{eq:neg_H}), this equals to minimizing the negative entropy, i.e., $\min_{\theta} -\mathcal{H}(\hat{\yy}_w)$. 
From now, we will use $\mathcal{L}_{\mathcal{H}}(\hat{\yy}_w)=\mathcal{H}(\hat{\yy}_w)$ to refer to the additional loss term computing the entropy for the misclassified pixels following Eq. \ref{eq:neg_H}. Note that given a uniform distribution $\qq$, maximizing the entropy of $\yy_w$ boils down to minimizing the \textit{Kullback-Leibler} (KL) divergence between $\yy_w$ and $\qq$. In what follows, we define another term based on this idea.

\vspace{0.2 cm}
\noindent \textbf{Proxy for entropy maximization:} In addition to explicitly maximizing the entropy of predictions (or to minimizing the negative entropy) as proposed in Eq. \ref{eq:neg_H}, we resort to an alternative regularizer, which is a variant of the KL divergence \cite{belharbi2020deep}. The idea is to encourage the output probabilities in $\yy_w$ (the misclassified pixels) to be close to the uniform distribution (i.e. all elements in the probability simplex vector $q$ are equal to $\frac{1}{K}$), resulting in max-uncertainty. This term is:

\begin{equation}
\mathcal{D}_{KL}(\qq||\hat \yy_w)\overset{\mathrm{K}}{=}\mathcal{H}(\qq,\hat \yy_w),
\label{eq:proxyKL}
\end{equation}

\noindent with $\qq$ being the uniform distribution and the symbol $\overset{\mathrm{K}}{=}$ representing equality up to an additive or multiplicative constant associated with the number of classes. We refer the reader to the Appendix I in \cite{belharbi2020deep} for the Proof of this KL divergence variant, as well as its gradients. It is important to note that despite both terms, \eqref{eq:neg_H} and \eqref{eq:proxyKL}, push $\hat \yy_w$ towards a uniform distribution, their gradient dynamics are different, and thus the effect on the weight updates differs. Here we perform an experimental analysis to assess which term leads to better performance. We will use $\mathcal{L}_{KL}(\hat{\yy}_w)=\mathcal{D}_{KL}(\qq||\hat \yy_w)$ to refer to the additional loss based on Eq. \ref{eq:proxyKL}.\\

\noindent \textbf{Global learning objective:}
Our final loss function takes the following form: 
$\mathcal{L}= \mathcal{L}_{Seg}(\yy,\hat{\yy}) - \lambda \mathcal{L}_{me}(\hat{\yy}_w)$,

where $\hat{\yy}$ is the entire set of pixel predictions, $\mathcal{L}_{Seg}$ the segmentation loss\footnote{$\mathcal{L}_{Seg}$ can take the form of any segmentation loss (e.g., CE or Dice)}, $\mathcal{L}_{me}$ is one of the proposed maximum entropy regularization terms and $\lambda$ balances the importance of each objective. Note that $\mathcal{L}_{me}$ can take the form of the standard entropy definition, i.e.  $\mathcal{L}_{me}(\hat{\yy}_w) = \mathcal{L}_{H}(\hat{\yy}_w)$ (eq. \eqref{eq:neg_H}) or the proxy for entropy maximization using the KL divergence, i.e. $\mathcal{L}_{me}(\hat{\yy}_w) = \mathcal{L}_{KL}(\hat{\yy}_w)$ (eq. \eqref{eq:proxyKL}). 
While the first term will account for producing good quality segmentations
the second term will penalize overconfident predictions only for challenging pixels, increasing the awareness of the model about the more uncertain image regions, maintaining high confidence in regions that are actually identified correctly.\\

\noindent \textbf{Baseline models:}

We trained baseline networks using a simple loss composed of a single segmentation objective $\mathcal{L}_{Seg}$, without adding any regularization term. 
We used the two most popular segmentation losses :
cross-entropy ($\mathcal{L}_{CE}$) and the negative soft Dice coefficient ($\mathcal{L}_{dice}$) as defined by \cite{milletari2016v}. Furthermore, we also compare our method to state-of-the-art calibration approaches.
First, due to its similarity with our work, we include the confidence penalty loss proposed in \cite{pereyra2017regularizing}, which discourages \textit{all} the neural network predictions from being over-confident by penalizing low-entropy distributions. This is achieved by adding a low-entropy penalty term over all the pixels (in contrast with our method that only penalizes the misclassified pixels), which can be defined as:
$\mathcal{L}_H(\hat{\yy})=-\frac{1}{|\hat{\yy}|} \sum_{k, i\in \hat{\yy} } \hat{p}_{i,k}\log \hat{p}_{i,k}.$

We train two baseline models using the aforementioned regularizer $\mathcal{L}_H(\hat{\yy})$, considering cross-entropy ($\mathcal{L}_{CE}$) and Dice losses ($\mathcal{L}_{dice}$). We also assess the performance of focal-loss \cite{lin2017focal}, since recent findings \cite{mukhoti2020calibrating} demonstrated the benefits of using this objective to train well-calibrated networks.

\noindent \textbf{Post-hoc calibration baselines.} We also included two well known calibration methods typically employed for classification \cite{guo2017calibration}: isotonic regression (IR) \cite{zadrozny2002transforming} and Platt scaling (PS) \cite{platt1999probabilistic}. 
Differently from our methods which only use the training split, IR and PS are trained using validation data \cite{guo2017calibration}, keeping the original network parameters fixed. This is an advantage of our approaches since they do not require to keep a hold-out set for calibration. We apply IR and PS to the predictions of the vanilla baseline models trained with $\mathcal{L}_{dice}$ and $\mathcal{L}_{CE}$ models.

\section{Experiments and results}
\noindent \textbf{Dataset and network details.}
We benchmark the proposed method in the context of Left Atrial (LA) cavity and White Matter Hyperintensities (WMH) segmentation in MR images. For LA, we used the Atrial Segmentation Challenge dataset \cite{xiong2020global}, which provides 100 3D gadolinium-enhanced MR imaging scans (GE-MRIs) and LA segmentation masks for training and validation. These scans have an isotropic resolution of 0.625 $\times$ 0.625 $\times$ 0.625mm$^3$. We used the splits and pre-processed data from \cite{yu2018pu} (80 scans for training and 20 for evaluation - 5\% of training images were used for validation). The WMH dataset \cite{kuijf2019standardized} consists of 60 MR images binary WMH masks. Each subject includes a co-registered 3D T1-weighted and a 2D multi-slice FLAIR of 1$\times$ 1$\times$3mm. We split the dataset into independent training (42), validation (3) and test (15) sets. 

We benchmark our proposed method with two state-of-the-art DNN architectures (UNet\cite{RonnebergerUnet15} and ResUNet \cite{zhang2018road}) implemented using Tensorflow 2.3
\footnote{Code: \url{ https://github.com/agosl/Maximum-Entropy-on-Erroneous-Predictions/}}
(results for ResUNet are included in the Supp. Mat.). During training, for the WMH dataset we extract patches of size $64\times64\times64$, and we train the networks until convergence by randomly sampling patches so that the central pixel corresponds to foreground label with 0.9 probability to account for label imbalance. For LA dataset all the scans were cropped to size $144\times144\times80$ and centered at the heart region. We used Adam optimizer with a batch size of 64 for WMH and 2 for LA. The learning rate was set to 0.0001, and reduced by a factor of 0.85 every 10 epochs. Hyper-parameters were chosen using the validation split, and results reported on the hold-out test set.\\

\noindent \textbf{Training details and evaluation metrics.}
 As baselines, we used networks trained with $\mathcal{L}_{CE}$ and  $\mathcal{L}_{dice}$ only. We also included the aforementioned post-hoc calibration methods (namely IR and PS) as post-processing step for these vanilla models.
 We also implemented the confidence penalty-based method \cite{pereyra2017regularizing} previously discussed by adding the entropy penalizer $\mathcal{L}_{H}(\hat{\yy})$ and using the hyper-parameter $\beta = 0.2$ suggested by the authors. We also include the focal-loss ($\mathcal{L}_{FL}$) with $\gamma=2$, following the authors' findings \cite{lin2017focal} and compare with the proposed regularizers which
 penalize low entropy in wrongly classified pixels: $\mathcal{L}_{H}(\hat{\yy}_w)$ (Eq. \ref{eq:neg_H}) and $\mathcal{L}_{KL}(\hat{\yy}_w)$ (Eq. \ref{eq:proxyKL}).
We performed grid search with different $\lambda$, and we found empirically that 0.3 works best for WMH models trained with $\mathcal{L}_{CE}$ and 1.0 for $\mathcal{L}_{dice}$. For the LA dataset, we chose 0.1 for $\mathcal{L}_{CE}$ and 0.5 for $\mathcal{L}_{dice}$. For each setting we trained 3 models and report the average results. 

To assess segmentation performance we resort to Dice Similarity Coefficient (DSC) and Hausdorff Distance (HD), whereas we use standard calibration metrics: %to assess the accuracy of the uncertainty estimates: %. We also included metrics affected by model calibration, namely the 
Brier score \cite{brier1950verification}, Stratified Brier score \cite{wallace2014improving} (adapted to image segmentation following \cite{larrazabal2021orthogonal}) and Expected Calibration Error \cite{naeini2015obtaining}. We also employ \textit{reliability diagrams}, depicting the observed frequency as a function of the class probability. Note that in a perfectly calibrated model, the frequency on each bin matches the confidence, and hence all the bars lie on the diagonal.\\

\begin{table}[t!]
\begin{center}
\begin{adjustbox}{width=1.0\textwidth}
\begin{tabular}{*{12}{c}}
\toprule
&&
\multicolumn{4}{c}{ {Segmentation performance}} & \multicolumn{6}{c}{Calibration performance} \\
\cmidrule(lr){3-6} \cmidrule(l){6-12}
\multicolumn{2}{c}{Training loss} &\multicolumn{2}{c}{Dice coefficient}
&\multicolumn{2}{c}{HD}
&\multicolumn{2}{c}{Brier ($10^{-4}$)}
&\multicolumn{2}{c}{ Brier$^{+}$}
&\multicolumn{2}{c}{ECE ($10^{-3}$)}\\
\cmidrule(lr){3-4}\cmidrule(lr){5-6} \cmidrule(l){7-8}\cmidrule(l){9-10}\cmidrule(l){11-12}
&&WMH&LA&WMH&LA&WMH&LA&WMH&LA&WMH&LA\\
\toprule
\multirow{5}{*}{$\mathcal{L}_{dice}$} & -- & \textbf{0.770 (0.100)}&\bf 0.886 (0.060)&24.041 (10.845)&28.282 (11.316)&6.717 (4.184)&29.182(15.068) &0.257 (0.125)&0.107 (0.090)&0.667 (0.414)&28.861 (15.009)\\

& + PS& 0.763 (0.103)&  0.884 (0.065)&24.151 (10.937)& \bf 26.565 (10.683)  &6.187 (3.974)&   24.953 (14.250)     &0.271 (0.126)&      0.114 (0.087)&1.563 (0.235)&16.346 (13.143)\\
& + IR&\textbf{0.770 (0.098)}&0.883 (0.065) &24.176 (10.725)& 26.699 (11.031)&5.541 (3.391)&24.617 (13.936) &\bf 0.212 (0.107)&0.111 (0.083) &1.539 (0.181) & 16.303 (13.670)\\

&$+\mathcal{L}_{H}(\hat{Y})$ \cite{pereyra2017regularizing}&0.769 (0.099)&0.885 (0.050)&21.608 (8.830)&29.811 (11.168)&6.751 (4.194)&29.019(12.709)&0.249 (0.125)&0.109 (0.077)&0.670 (0.415)&28.458 (12.514)\\

 &   \cellcolor{LightGray} $+\mathcal{L}_{H}(\hat{Y}_w)$&  \cellcolor{LightGray} 0.758 (0.108)&  \cellcolor{LightGray} 0.873 (0.069)&  \cellcolor{LightGray} 21.243 (8.755)&  \cellcolor{LightGray} 29.374 (10.965)&  \cellcolor{LightGray} 5.874 (3.875)&  \cellcolor{LightGray} 24.709(13.774)&  \cellcolor{LightGray} 0.244 (0.124)&  \cellcolor{LightGray} 0.103 (0.086)&
 \cellcolor{LightGray} 0.510 (0.350)&
 \cellcolor{LightGray} 18.796 (15.005)\\

& \cellcolor{LightGray} $+\mathcal{L}_{KL}(\hat{Y}_w)$&  \cellcolor{LightGray} \textbf{0.770 (0.098)}&  \cellcolor{LightGray} 0.881 (0.064)&  \cellcolor{LightGray} \textbf{20.804 (8.122)}&  \cellcolor{LightGray} 28.415 (12.860)&  \cellcolor{LightGray} \textbf{5.564 (3.586)}&  \cellcolor{LightGray} \textbf{23.182(12.464)}&  \cellcolor{LightGray} 0.231 (0.114)&  \cellcolor{LightGray} \textbf{0.095 (0.077)}&  \cellcolor{LightGray} \textbf{0.471 (0.318)}& \cellcolor{LightGray} \textbf{15.587 (13.391)}\\

%& $\mathcal{L}_{dice}$ + IR & 0.770 (0.098)& &24.176 (10.725)& &0.212 (0.107)&& 1.539 (0.181)&&5.541 (3.391)\\

\midrule
\multirow{4}{*}{$\mathcal{L}_{CE}$} & -- &0.755 (0.111)&0.878 (0.070)&21.236 (7.735)&\textbf{27.163 (11.967)}&6.462 (4.141)&24.447 (14.876)&0.280 (0.140)&0.108 (0.092)&0.620 (0.400)&18.383 (16.700)\\

& + PS& 0.763 (0.105)&  0.878 (0.069)  &21.008 (7.637)& 27.203 (11.963)   &5.459 (3.367)&  23.458 (13.462) &0.214 (0.115)& 0.100 (0.081) &1.631 (0.188) & 16.576 (15.427)\\

& + IR &0.764 (0.105)& 0.878 (0.070)&21.202 (7.855)&27.223 (11.944) &5.430 (3.326)& 23.544 (13.803) & \bf 0.210 (0.112)& 0.102 (0.084)&1.622 (0.198)&16.421 (15.500)\\

&$+\mathcal{L}_{H}(\hat{Y})$ \cite{pereyra2017regularizing}&0.760 (0.109)&0.881 (0.070)&23.124 (9.523)&29.464 (14.389)&6.369 (4.018)&23.539 (11.903)&0.242 (0.125)&0.096 (0.070)&4.100 (0.582)&15.590 (14.002)\\

& \cellcolor{LightGray} $+\mathcal{L}_{H}(\hat{Y}_w)$&  \cellcolor{LightGray} 0.770 (0.095)& \cellcolor{LightGray} \textbf{0.883 (0.058)}& \cellcolor{LightGray} \textbf{19.544 (7.254)}&  \cellcolor{LightGray}28.560(13.352)& \cellcolor{LightGray}5.417 (3.547)& \cellcolor{LightGray}\textbf{22.506 (11.903)}& \cellcolor{LightGray}0.217 (0.104)&  \cellcolor{LightGray}\bf 0.093 (0.071) &  \cellcolor{LightGray}0.436 (0.301)&  \cellcolor{LightGray} \bf 15.242 (13.730)\\

&  \cellcolor{LightGray} $+\mathcal{L}_{KL}(\hat{Y}_w)$& \cellcolor{LightGray} \textbf{0.777 (0.093)}& \cellcolor{LightGray}0.876 (0.070)& \cellcolor{LightGray}22.298 (9.566)& \cellcolor{LightGray}28.736 (11.972)& \cellcolor{LightGray}\textbf{5.331 (3.478)}& \cellcolor{LightGray}24.085 (13.330)& \cellcolor{LightGray}0.213 (0.099)& \cellcolor{LightGray} 0.105 (0.090)& \cellcolor{LightGray}\textbf{0.422 (0.289)}& \cellcolor{LightGray}17.348 (14.786)\\

%& $\mathcal{L}_{CE}$ + IR&0.764 (0.105)& &21.202 (7.855)& &0.210 (0.112)& &1.622 (0.198)& &5.430 (3.326) \\

%hasta aca
%aca fl
\midrule
\multicolumn{2}{c}{$\mathcal{L}_{FL}$}&0.753 (0.113)&0.881 (0.064)&21.931 (8.167)&28.599 (11.968)&5.760 (3.732)& 23.928 (11.626)&0.243 (0.130)&0.095 (0.066)&0.438 (0.310)&25.998 (12.740)\\
\bottomrule
\end{tabular}
\end{adjustbox}
\end{center}
    \caption{Mean accuracy and standard deviation for both WMH and LS segmentation tasks with UNet as backbone. Our models are gray-shadowed and best results are highlighted in bold.}
    \vspace{-3 mm}%\textcolor{red}{Jose: Esta tabla es con solo UNet..}}
\label{tab:results_UNet}
\end{table}

%\noindent \textbf{Experimental results.}
\noindent \textbf{Results.} Our main goal is to improve the estimated uncertainty of the predictions, while retaining the segmentation power of original losses. Thus, we first assess whether integrating our regularizers leads to a performance degradation. Table \ref{tab:results_UNet} reports the results across the different datasets with the UNet model (results for ResUNet are included in the Supp. Mat.). First, we can observe that adding the proposed regularizers does not result in a remarkable loss of segmentation performance. Indeed, in some cases, e.g., $\mathcal{L}_{dice} + \mathcal{L}_{KL}(\hat{Y}_w)$ in WMH, the proposed model outperforms the baseline by more than 3\% in terms of HD. Furthermore, this behaviour holds when the standard CE loss is used in conjunction with the proposed terms, suggesting that the overall segmentation performance is not negatively impacted by adding our regularizers into the main learning objective. Last, it is noteworthy to mention that even though $\mathcal{L_{FL}}$ sometimes outperforms the baselines, it typically falls behind our two losses. In terms of qualitative performance, Fig. \ref{fig:qualitative_LA}.a depicts exemplar cases of the improvement in probability maps obtained for each loss function in LA segmentation.

%\noindent \textbf{Calibration performance.}
Regarding calibration performance, 
%Our main focus in this section is to evaluate the calibration quality of a network that is trained with different losses. R
recent empirical evidence \cite{mehrtash2020confidence} shows that, despite leading to strong predictive models, CE and specially Dice losses result in highly-confident predictions. The results obtained for calibration metrics (Brier and ECE in Table \ref{tab:results_UNet}) are in line with these observations. These results evidence that regardless of the dataset, networks trained with any of these losses as a single objective, lead to worse calibrated models compared to the proposed penalizers. Explicitly penalizing low-entropy predictions over all the pixels, as in \cite{pereyra2017regularizing}, typically improves calibration. Nevertheless, despite the gains observed with \cite{pereyra2017regularizing}, empirical results demonstrate that penalizing low-entropy values \textit{only over misclassified pixels} brings the largest improvements, regardless of the main segmentation loss used. In particular, the proposed MEEP regularizers outperform the baselines in all the three calibration metrics and in both datasets, with improvements ranging from 1\% to 13\%, except for Brier+ in WMH. However, in this case, even though IR achieves a better Brier$^+$, it results in worse ECE.

%%--- Calibration curves
\begin{figure} [t!]
     \begin{subfigure}[b]{0.24\textwidth}
         \centering
         \includegraphics[width=0.95\linewidth]{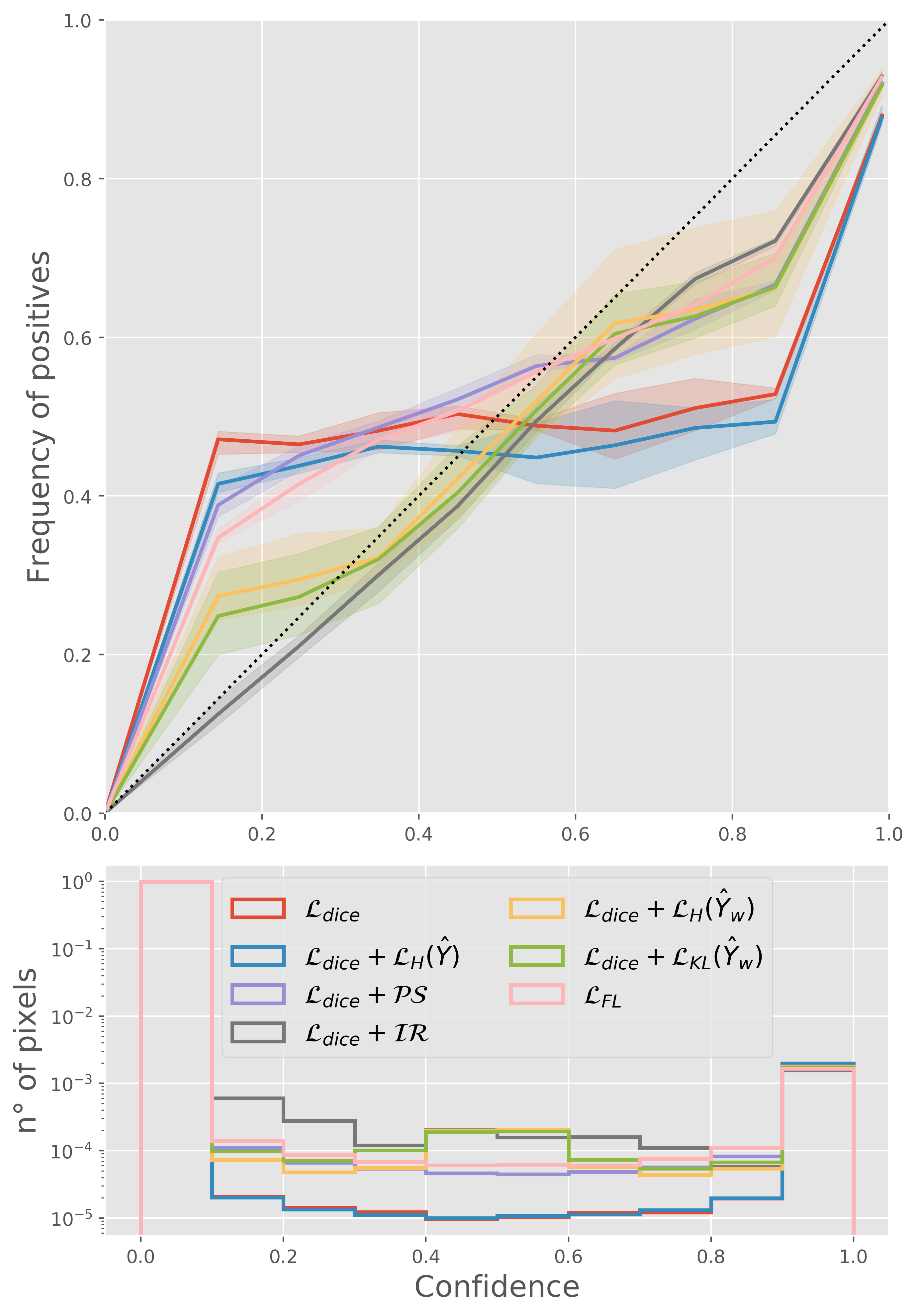}
         \caption{WMH}
     \end{subfigure}
     \hfill
     \begin{subfigure}[b]{0.24\textwidth}
         \centering
         \includegraphics[width=0.95\linewidth]{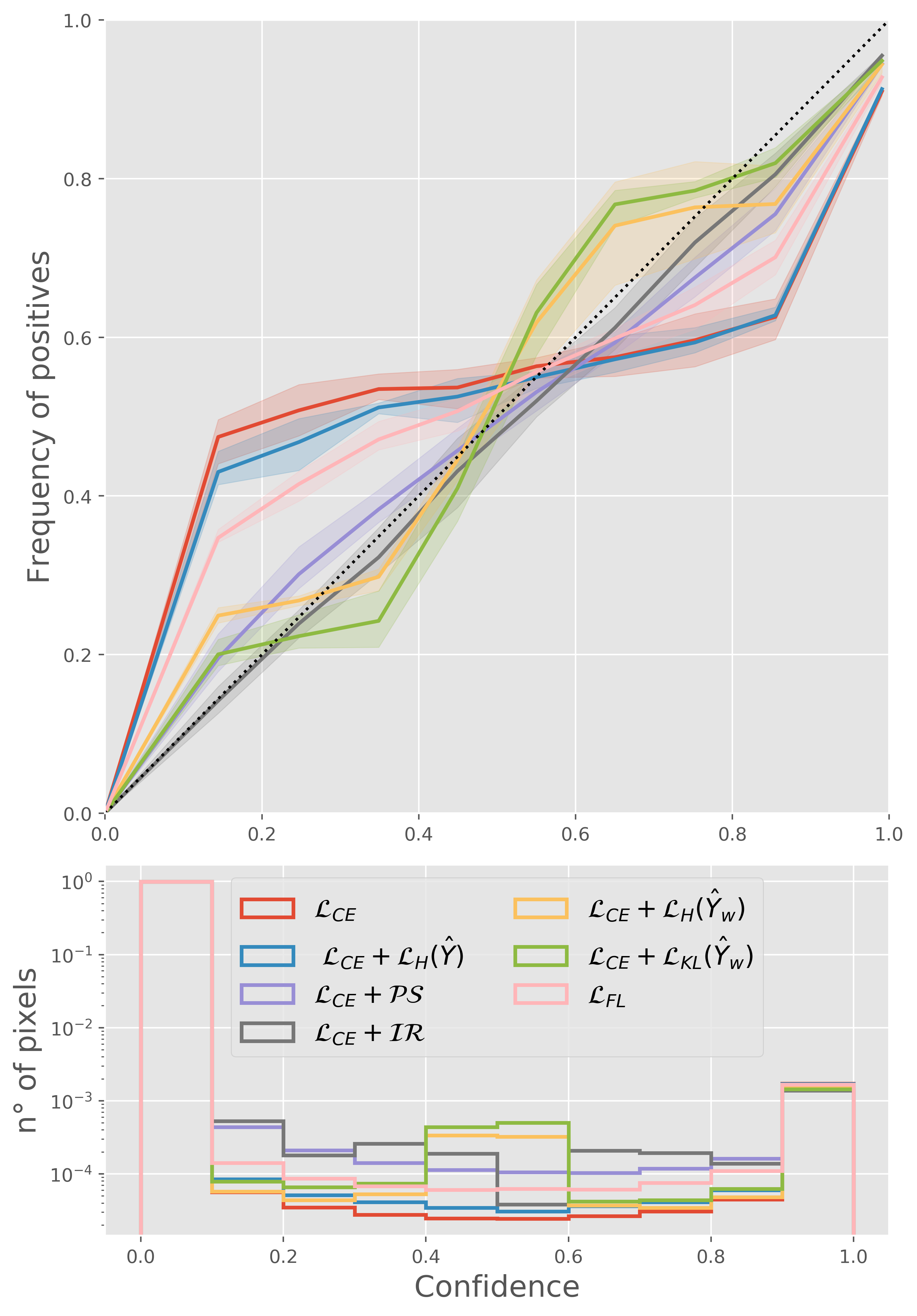}
          \caption{WMH}
     \end{subfigure}
     \hfill
     \begin{subfigure}[b]{0.24\textwidth}
         \centering
         \includegraphics[width=0.95\linewidth]{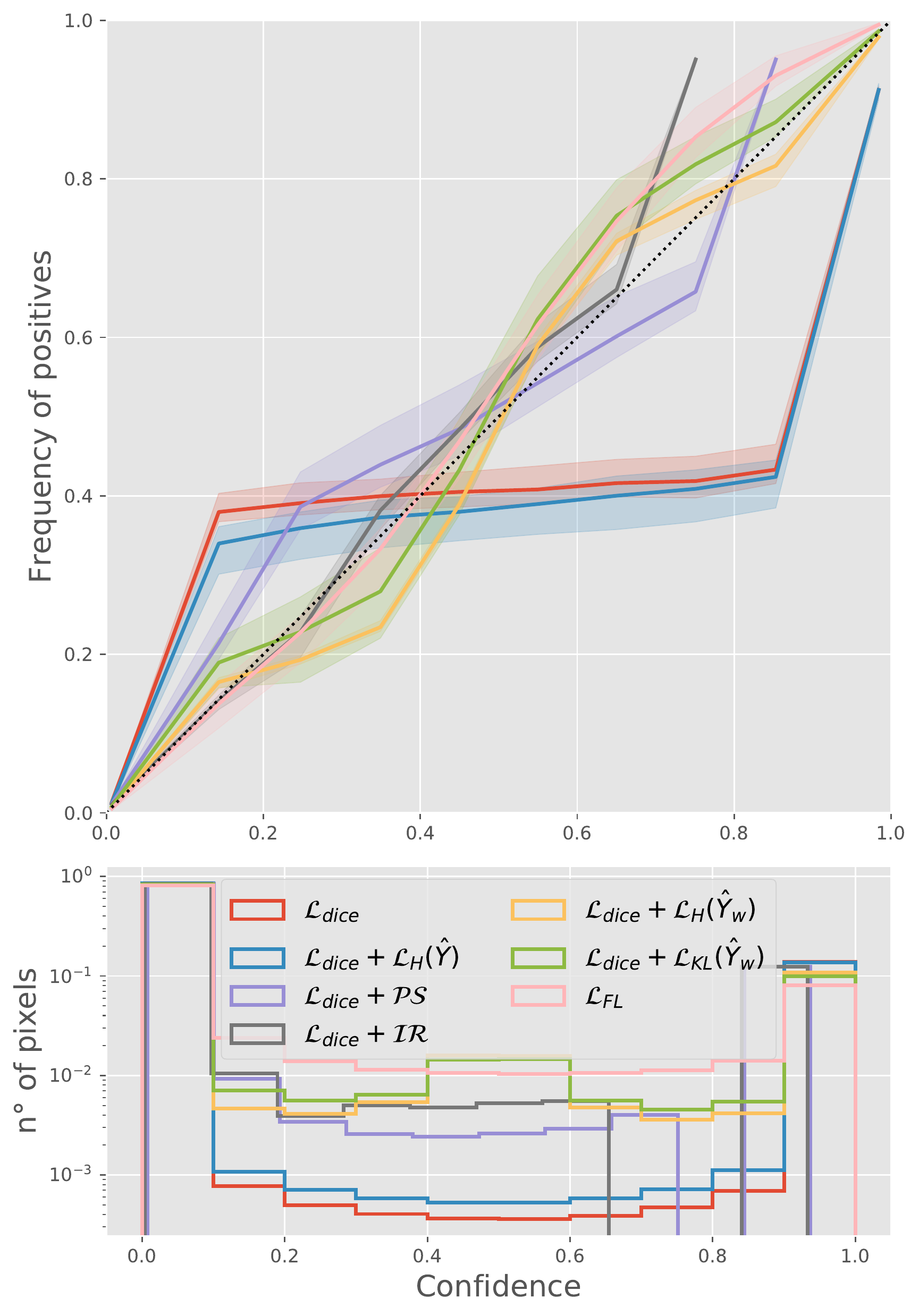}
          \caption{LA}
     \end{subfigure}
     \hfill
     \begin{subfigure}[b]{0.24\textwidth}
         \centering
         \includegraphics[width=0.95\linewidth]{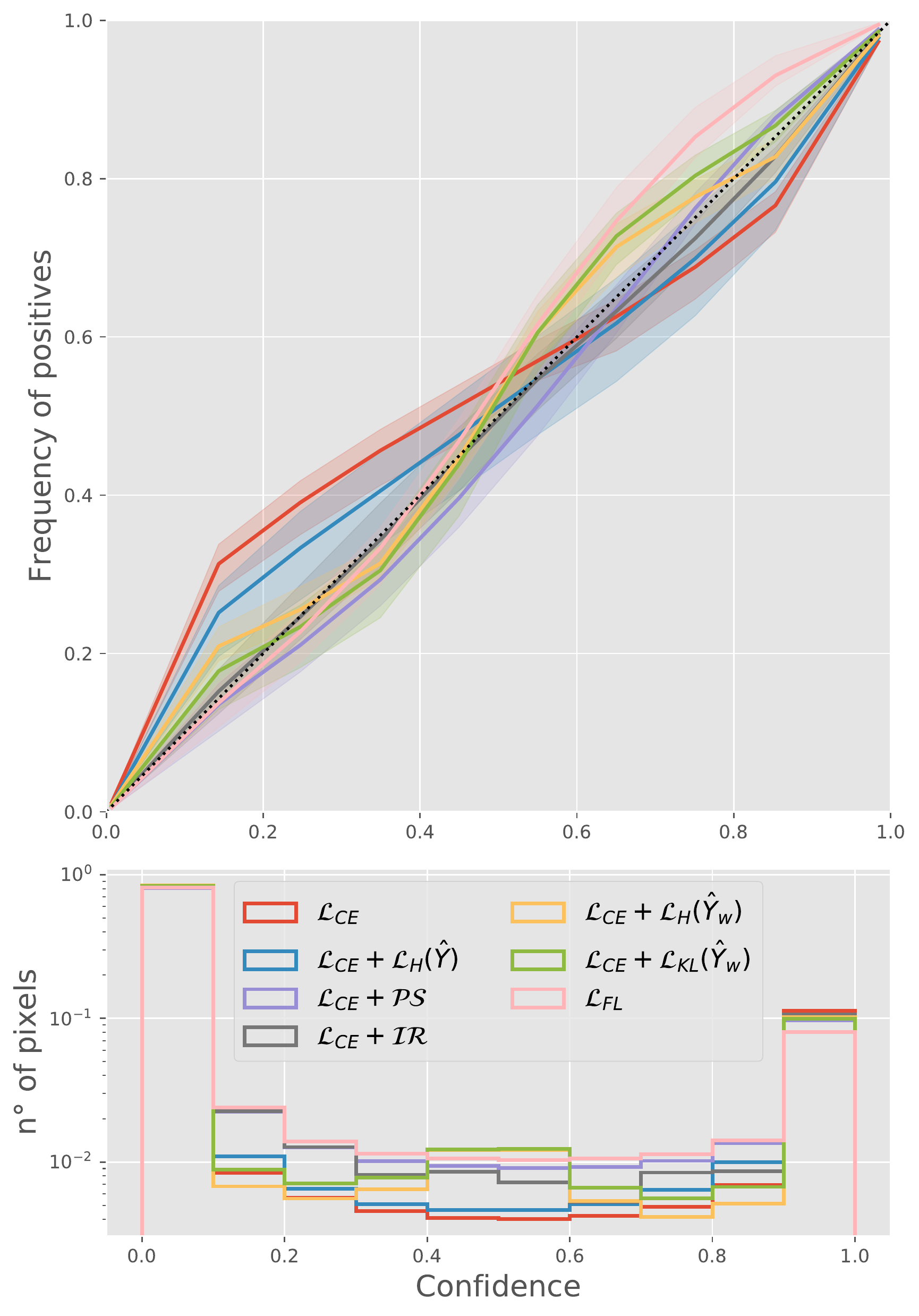}
          \caption{LA}
     \end{subfigure}

  \caption{The first row shows the reliability plot calculated on the entire volume of test images for each of the models while the bottom row shows histogram of probabilities produced by each method. }

\label{fig:calibration_curve_LV}
\end{figure}
When evaluating the proposed MEEP regularizers ($\mathcal{L}_{KL}(\hat{Y}_w)$ and $\mathcal{L}_{H}(\hat{Y}_w)$) combined with the segmentation losses based on DSC and CE, we observe that DSC with $\mathcal{L}_{KL}(\hat{Y}_w)$ consistently achieves better performance in most of the cases. However, for CE, both regularizers alternate best results, which depend on the dataset used. We hypothesize that this might be due to the different gradient dynamics shown by the two regularizers\footnote{We refer to Fig 3 and Appendix I in \cite{belharbi2020deep} for a detailed explanation regarding the different energies for binary classification and their derivatives.}. 

Regarding the focal loss, even though it improves model calibration when compared with the vanilla models, we observe that the proposed regularizers 
\section{Conclusions}
In this paper, we presented a simple yet effective approach to improve the uncertainty estimates inferred from segmentation models when trained with popular segmentation losses. In contrast to prior literature, our regularizers penalize high-confident predictions only on misclassified pixels, increasing network uncertainty in complex scenarios. In addition to directly maximizing the entropy on the set of erroneous pixels, we present a proxy for this term, formulated with a KL regularizer modeling high uncertainty over those pixels. Comprehensive results on two popular datasets, losses and architectures demonstrate the potential of our approach. Nevertheless, we have also identified several limitations. For example, we have not assessed the effect of the proposed regularizers under severe domain shift (e.g. when testing on images of different organs). In this case it is not clear whether the model will output highly uncertain posteriors, or result again on overconfident but wrong predictions.  

\section{Supplementary Material}

\begin{table*}[h]
\begin{center}
\begin{adjustbox}{width=1.0\textwidth}
\begin{tabular}{*{12}{c}}
\toprule
&&
\multicolumn{4}{c}{ {Segmentation performance}} & \multicolumn{6}{c}{Calibration performance} \\
\cmidrule(lr){3-6} \cmidrule(l){6-12}
\multicolumn{2}{c}{Training loss} &\multicolumn{2}{c}{Dice coefficient}
&\multicolumn{2}{c}{HD}
&\multicolumn{2}{c}{Brier ($10^{-4}$)}
&\multicolumn{2}{c}{ Brier$^{+}$}
&\multicolumn{2}{c}{ECE ($10^{-3}$)}\\
\cmidrule(lr){3-4}\cmidrule(lr){5-6} \cmidrule(l){7-8}\cmidrule(l){9-10}\cmidrule(l){11-12}
&&WMH&LA&WMH&LA&WMH&LA&WMH&LA&WMH&LA\\
\toprule
\multirow{4}{*}{$\mathcal{L}_{dice}$} & -- & 0.768 (0.108)&\bf 0.905 (0.024)&20.499 (8.468)&24.311 (7.733)&6.742 (4.346)&24.940 (7.692)&0.258 (0.136)&0.092 (0.043)&0.667 (0.429) & 24.698 (7.606)\\

&+ PS& 0.764 (0.110)& 0.904 (0.026)&20.553 (8.313)& 23.412 (7.116)&6.120 (4.010) & 21.248 (6.916) &0.253 (0.132)& 0.099 (0.042) & 1.663 (0.256) & 10.933 (4.285)\\
&+ IR&0.770 (0.105)& 0.904 (0.026) &20.518 (8.470)& \bf 23.038 (6.535) &5.616 (3.512)& 21.034 (6.866) &0.194 (0.111)& 0.097 (0.041) &1.605 (0.263) & \bf 10.319 (5.483) \\

&$+\mathcal{L}_{H}(\hat{Y}$)&0.754 (0.122)&0.903 (0.029)&21.089 (7.475)&23.811 (8.902)&7.013 (4.643)&24.952 (8.964)&0.267 (0.157)&0.086 (0.051)&0.696 (0.461) & 24.457 (8.765)\\

& \cellcolor{LightGray} $+\mathcal{L}_{H}(\hat{Y}_w$)& \cellcolor{LightGray}\textbf{0.786 (0.089)}& \cellcolor{LightGray}0.903 (0.025)& \cellcolor{LightGray}20.033 (7.566)& \cellcolor{LightGray}24.095 (8.357)& \cellcolor{LightGray}5.451 (3.492)& \cellcolor{LightGray}\bf 19.565 (6.493)& \cellcolor{LightGray}0.183 (0.095)& \cellcolor{LightGray}0.083 (0.036)& \cellcolor{LightGray}0.451 (0.287)&  \cellcolor{LightGray}13.006 (5.160)\\

& \cellcolor{LightGray} $+\mathcal{L}_{KL}(\hat{Y}_w$)& \cellcolor{LightGray} \textbf{0.786 (0.093)}& \cellcolor{LightGray}0.900 (0.028)& \cellcolor{LightGray}\textbf{18.848 (6.513)}& \cellcolor{LightGray}23.600 (6.496)& \cellcolor{LightGray}\textbf{5.379 (3.502)}& \cellcolor{LightGray}20.106 (6.971)& \cellcolor{LightGray}\textbf{0.174 (0.094)}& \cellcolor{LightGray}\textbf{0.079 (0.042)}& \cellcolor{LightGray}\textbf{0.434 (0.285)}&  \cellcolor{LightGray} 12.561 (6.222)\\

\midrule

\multirow{4}{*}{$\mathcal{L}_{CE}$} & -- &0.770 (0.104)&0.890 (0.035)&18.928 (7.175)&26.596 (8.121)&6.256 (4.044)&22.458 (8.417)&0.259 (0.128)&0.113 (0.055)&0.602 (0.390)& 16.700 (8.763)\\

& + PS& 0.775 (0.101)&0.893 (0.033) &19.385 (7.667)& 25.508 (8.387) &5.296 (3.324)&21.092 (6.955) &0.194 (0.101)& 0.088 (0.044) &1.637 (0.168) & 12.140 (5.482)\\
& + IR&   0.775 (0.100)& 0.892 (0.034)  &19.649 (7.834)& 26.041 (8.314) &5.236 (3.229)& 21.023 (7.160) &\bf 0.184 (0.096)& \bf 0.091 (0.045) &1.590 (0.183) & \bf 11.765 (5.758)\\

&$+\mathcal{L}_{H}(\hat{Y}$)&0.778 (0.092)&\bf 0.896 (0.030)&\textbf{18.554 (7.214)}&\bf 25.137 (8.291)&6.208 (3.842)&20.933 (7.552)&0.227 (0.103)&0.098 (0.044)&5.251 (0.504)& 11.784 (6.481)\\

& \cellcolor{LightGray} $+\mathcal{L}_{H}(\hat{Y}_w$)& \cellcolor{LightGray}\textbf{0.779 (0.096)}& \cellcolor{LightGray}0.890 (0.036)& \cellcolor{LightGray}18.789 (7.205)& \cellcolor{LightGray}25.349 (6.265)& \cellcolor{LightGray}\textbf{5.169 (3.347)}& \cellcolor{LightGray}21.721 (7.639)& \cellcolor{LightGray}0.191 (0.093)& \cellcolor{LightGray}0.106 (0.053)& \cellcolor{LightGray}0.396 (0.257)&  \cellcolor{LightGray}14.607 (7.872)\\

& \cellcolor{LightGray} $+\mathcal{L}_{KL}(\hat{Y}_w$)&  \cellcolor{LightGray}0.775 (0.095)& \cellcolor{LightGray}0.895 (0.032)& \cellcolor{LightGray}20.949 (9.769)& \cellcolor{LightGray}25.576 (7.426)& \cellcolor{LightGray}5.269 (3.383)& \cellcolor{LightGray}\textbf{20.609 (7.260)}& \cellcolor{LightGray}0.187 (0.091)& \cellcolor{LightGray}0.097 (0.047)& \cellcolor{LightGray}\textbf{0.390 (0.248)}&  \cellcolor{LightGray}12.811 (7.250)\\

%hasta aca

%aca fl
\midrule
\multicolumn{2}{c}{$\mathcal{L}_{FL}$}&0.780 (0.090)&0.891 (0.031)&19.759 (7.372)&26.447 (7.442)&5.621 (3.703)&21.269 (6.681)&0.216 (0.103)&0.102 (0.041)&0.472 (0.327) & 14.249 (6.140)\\
\bottomrule
\end{tabular}
\end{adjustbox}
\end{center}
    \caption{\textbf{Is our method backbone-agnostic?} This table includes the results for an additional experiment for both WMH and LA segmentation, using the ResUNet architecture as backbone (instead of UNet as used in the main manuscript), to show that our method is independent of the architecture. Our models are gray-shadowed and best results are highlighted in bold. Results show that the proposed regularizers typically lead to improvement on both model calibration and segmentation also for ResUNet architecture. This improvement is further stressed for the calibration metrics, where the four variants of our method significantly outperform the rest for WMH.} %\textcolor{red}{Jose: Esta tabla es con solo ResNet..}}
\label{tab:results_ResNet}
\end{table*}

 %by a wide margin. %Thus, these results evidence that coupling the proposed terms with the main segmentation loss yields better calibrated networks without sacrificing discriminative power.
%WMH and LA segmentation are presented in Table \ref{tab:results_WMH} and Table \ref{tab:results_LA}, respectively. 

\begin{figure*}[t!]
     \centering
     \begin{subfigure}[b]{0.98\textwidth}
         \centering
         \includegraphics[width=\textwidth]{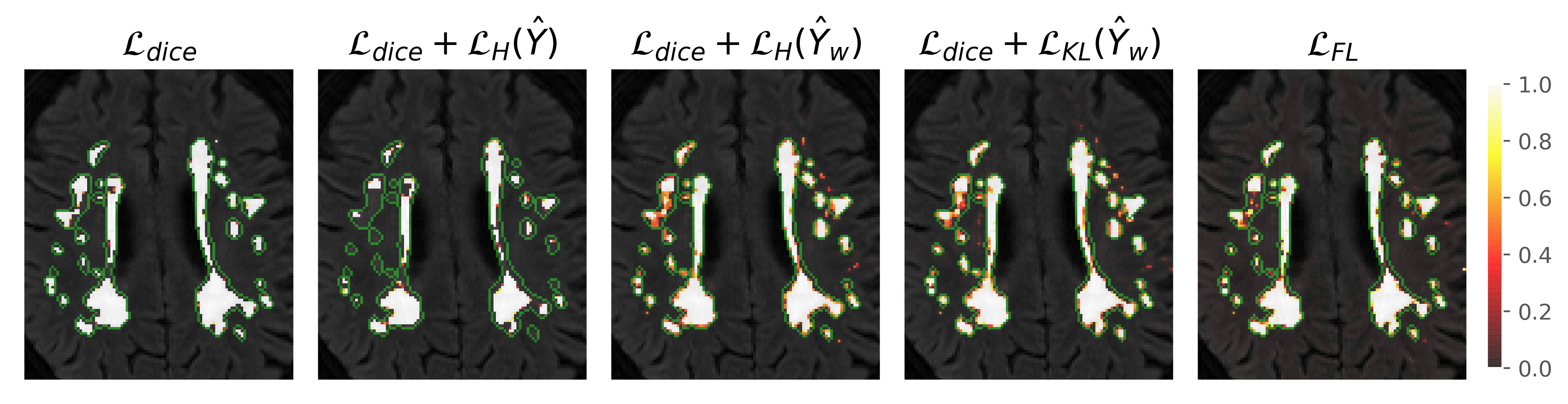}
     \end{subfigure}
     \hfill
    %  \begin{subfigure}[b]{0.98\textwidth}
    %      \includegraphics[width=0.95\textwidth]{images/dice_models50cc.png}
       
    %   \end{subfigure}
    %       \centering
     \begin{subfigure}[b]{0.98\textwidth}
         \centering
         \includegraphics[width=\textwidth]{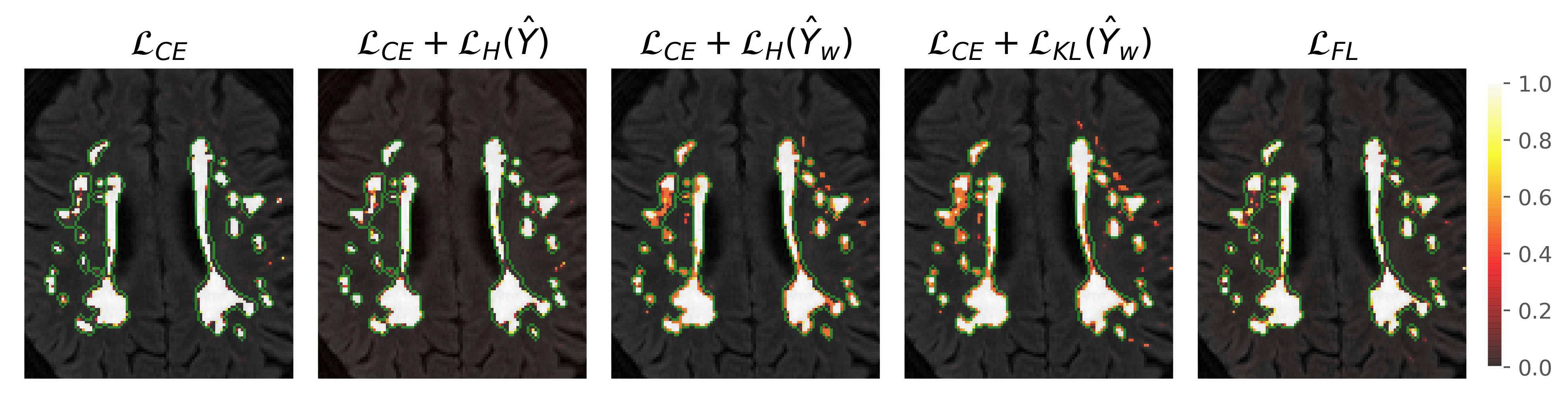}
     \end{subfigure}
    %   \hfill
    %   \begin{subfigure}[b]{0.98\textwidth}
         
    %     \includegraphics[width=0.95\textwidth]{images/CE_models50cc.png}
       
    % \end{subfigure}
 
     \caption{\textbf{Qualitative results for WMH segmentation} We show exemplar cases of the probability maps obtained for each loss function for the WMH segmentation task. As it can be observed, the predictions made by the vanilla network trained with DSC loss tend to be highly overconfident, either assigning probabilities equal to 0 or 1. However, when employing the proposed regularizers, the models tend to use the full range of possible values, assigning scores around 0.5 (marked in red) to the more challenging pixels. }
       
 \label{fig:qualitative_WMH}
\end{figure*}

\section{Acknowledgments}

The authors gratefully acknowledge NVIDIA Corporation with the donation of the GPUs used for this research, the support of UNL with the CAID program and ANPCyT (PRH-2019-00009). EF is supported by the Google Award for Inclusion Research (AIR) Program. AL was partiallly supported by the Emerging Leaders in the Americas Program (ELAP) program. We also thank Calcul Quebec and Compute Canada.

\bibliographystyle{splncs}
\bibliography{bibliography.bib}

\end{document}